\documentclass[nohyperref]{article}

\usepackage{microtype}
\usepackage{graphicx}
\usepackage{booktabs} 

\usepackage{multirow}
\usepackage{subcaption}
\usepackage{caption}
\usepackage{tabularx} 
\usepackage{adjustbox}
\usepackage{amsmath,amssymb}
\usepackage[table,xcdraw]{xcolor}
\usepackage{tikz}
\usepackage{lipsum}

\usepackage{ctable} 

\usepackage{hyperref}

\usepackage[accepted]{icml2022}

\usepackage{amsmath}
\usepackage{amssymb}
\usepackage{mathtools}
\usepackage{amsthm}

\usepackage[capitalize,noabbrev]{cleveref}

\theoremstyle{plain}

\theoremstyle{definition}

\theoremstyle{remark}

\usepackage[textsize=tiny]{todonotes}

\icmltitlerunning{Exploring Adversarial Attacks and Defenses in Vision Transformers trained with DINO}

\begin{document}

\twocolumn[
\icmltitle{Exploring Adversarial Attacks and Defenses in \\
Vision Transformers trained with DINO}



\icmlsetsymbol{equal}{*}

\begin{icmlauthorlist}
\icmlauthor{Javier Rando}{equal,eth}
\icmlauthor{Nasib Naimi}{equal,eth}
\icmlauthor{Thomas Baumann}{equal,eth}
\icmlauthor{Max Mathys}{eth}
\end{icmlauthorlist}

\icmlaffiliation{eth}{ETH Zurich, Switzerland}

\icmlcorrespondingauthor{Javier Rando}{jrando@student.ethz.ch}
\icmlcorrespondingauthor{Nasib Naimi}{nnaimi@student.ethz.ch}
\icmlcorrespondingauthor{Thomas Baumann}{thobauma@student.ethz.ch}

\icmlkeywords{Machine Learning, Adversarial Attacks, Vision Transformers, ICML}

\vskip 0.3in
]



\printAffiliationsAndNotice{\icmlEqualContribution} 

\begin{abstract}
This work conducts the first analysis on the robustness against adversarial attacks on self-supervised Vision Transformers trained using DINO. First, we evaluate whether features learned through self-supervision are more robust to adversarial attacks than those emerging from supervised learning. Then, we present properties arising for attacks in the latent space. Finally, we evaluate whether three well-known defense strategies can increase adversarial robustness in downstream tasks by only fine-tuning the classification head to provide robustness even in view of limited compute resources. These defense strategies are: Adversarial Training, Ensemble Adversarial Training and Ensemble of Specialized Networks.
\end{abstract}

\section{Introduction}



Adversarial attacks \cite{szegedy2013intriguing} are one of the major challenges faced by current machine learning models. These attacks perturb inputs to provoke incorrect predictions while preserving input similarity, even on state-of-the-art architectures \cite{goodfellow2014explaining}. They are especially dangerous in safety-critical applications. Effectively defending against them is an open problem that seems to be in odds with the accuracy of the models \cite{roth2019odds}.

Meanwhile, advances in Computer Vision result in new architectures that are able to increase accuracy under existing benchmarks. Recently, Vision Transformers (ViTs) \cite{kolesnikov2021image} were presented as a successful application to vision of the foundational Transformer \cite{vaswani2017attention} used extensively in Natural Language Processing (NLP) tasks. However, this first ViT architecture required labelled data for training, imposing an important bottleneck for scalability. Caron et al. introduced a successful self-supervised training scheme, DINO \cite{caron2021emerging}. The authors claim that self-attention in models trained using DINO aligns with human interpretable features.

This motivates our evaluation of whether the representation learnt through DINO yields increased robustness against adversarial attacks that preserve input similarity. This work analyzes for the first time the effect that adversarial attacks have on self-supervised Vision Transformers trained using DINO. We first show that these ViTs are vulnerable in light of different adversarial attacks. Then, we conduct an in-depth analysis of how the attacks perturb the latent space generated by the Transformer encoder to provoke misclassification. Using these insights, we discuss potential efficient defense strategies that do not require retraining the Transformer backbone in scenarios where compute resources are limited. Overall, our contributions are as follow\footnote{Our code can be accessed on GitHub: \href{https://github.com/thobauma/AADefDINO}{https://github.com/thobauma/AADefDINO}}:
\vspace{-0.8em}
\begin{itemize}
    \item First evaluation of self-supervised DINO ViTs against different adversarial attacks, and analysis of properties emerging in their latent space. We also show that self-supervision increases attack transferability between ViTs and convolutional networks.
    \item Drawing on the mentioned properties, we motivate and empirically validate the suitability of the latent space of ViTs for precise detection and classification of adversarial inputs.
    \item Comparison of three defense strategies under the assumption of limited compute resources: Adversarial Training, Ensemble Adversarial Training and Ensemble of Specialized Networks.
\end{itemize}
\vspace{-0.8em}

\section{Related Work}

\subsection{Vision Transformers}

Transformers \cite{vaswani2017attention} were first presented in the field of Natural Language Processing (NLP) as a novel architecture relying entirely on self-attention mechanisms. They can be efficiently pre-trained on unlabeled data allowing for extracting knowledge from huge datasets without human annotations, e.g. web corpora. The resulting knowledge is then leveraged to solve specific downstream tasks by means of supervised fine-tuning \cite{brown2020language}. The same architecture was then transferred to Computer Vision. In a first attempt, supervised learning was presented as the only way to build so called Vision Transformers \cite{kolesnikov2021image}. Although they outperformed state-of-the-art architectures in most vision tasks \cite{khan2021transformers}, they required huge amounts of labeled data and computational power. Later, a successful self-supervised training framework, DINO, was presented to avoid the need for labeled data and exploit the advantages that turned Transformers into the leading architecture for NLP \cite{caron2021emerging}. Moreover, experimental results show that self-attention of Transformers trained using DINO contain explicit information about semantic segmentation of an image and align with human interpretable features. After DINO, other self-supervised training frameworks have been proposed such as Masked Autoencoders \cite{he2021masked} and BEiT \cite{bao2021beit}.

Different architectures have been proposed for Vision Transformers by tuning the number of parameters, or by creating "hybrid" models combining Transformers and CNNs \cite{kolesnikov2021image}. In this work, we will only consider ViTs relying completely on self-attention, and follow the nomenclature provided by Caron et al., i.e. ViT-S/16 refers to the "small" (S) network size (21M parameters) with input patches of size 16x16 \cite{caron2021emerging}.

\begin{table*}[!h]
\centering
\begin{adjustbox}{max width=\textwidth}
\begin{tabular}{|c|ccc|ccc|c|}
\hline
\multirow{2}{*}{\textbf{Clean}} & \multicolumn{3}{c|}{\textbf{FGSM}}                                                  & \multicolumn{3}{c|}{\textbf{PGD}}                                                  & \textbf{C\&W}           \\ \cline{2-8} 
                                & \multicolumn{1}{c|}{$\epsilon = 0.001$}  & \multicolumn{1}{c|}{$\epsilon = 0.03$}   & {$\epsilon = 0.1$}   & \multicolumn{1}{c|}{$\epsilon = 0.001$}  & \multicolumn{1}{c|}{$\epsilon = 0.03$}  & {$\epsilon = 0.1$}  & {$c = 50$}                      \\ \hline
\includegraphics[width=0.1\textwidth]{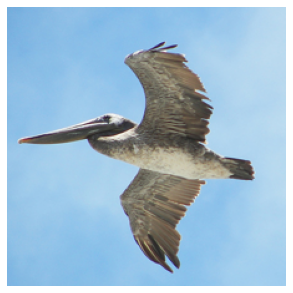}                            & \multicolumn{1}{l|}{\includegraphics[width=0.1\textwidth]{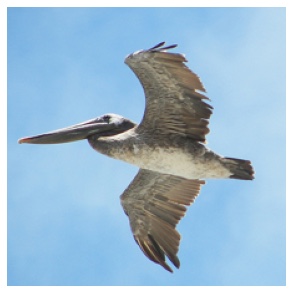}} & \multicolumn{1}{l|}{\includegraphics[width=0.1\textwidth]{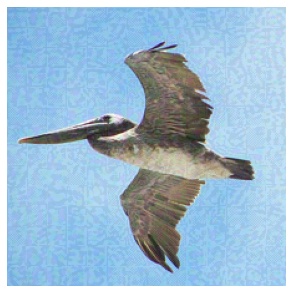}} & \includegraphics[width=0.1\textwidth]{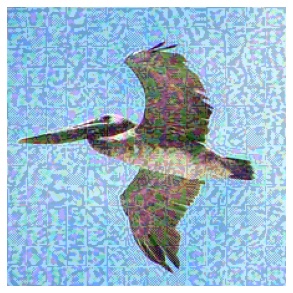}                     & \multicolumn{1}{l|}{\includegraphics[width=0.1\textwidth]{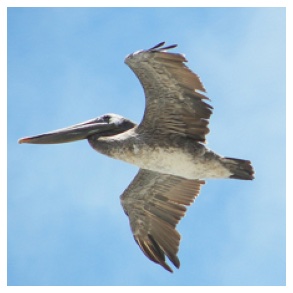}}  & \multicolumn{1}{l|}{\includegraphics[width=0.1\textwidth]{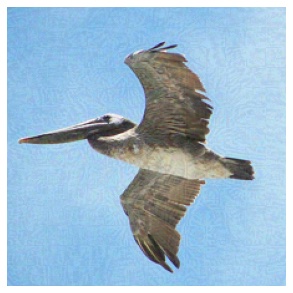}} & \includegraphics[width=0.1\textwidth]{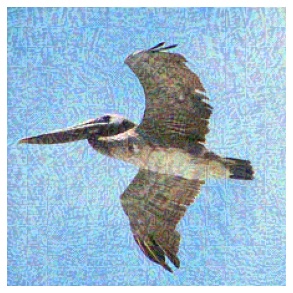}                       & \multicolumn{1}{l|}{\includegraphics[width=0.1\textwidth]{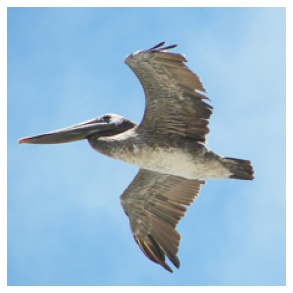} } \\ \hline\hline
Pelican  & \multicolumn{1}{c|}{Pelican} & \multicolumn{1}{c|}{{\color[HTML]{AB1E00} Kite}} & {\color[HTML]{AB1E00} Kite}                    & \multicolumn{1}{c|}{Pelican}  & \multicolumn{1}{c|}{{\color[HTML]{AB1E00} Kite}} & {\color[HTML]{AB1E00} Kite}                     & \multicolumn{1}{c|}{{\color[HTML]{AB1E00} Albatross}} \\ \hline
\end{tabular}
\end{adjustbox}
\caption{Adversarial perturbations generated on a validation sample using each of the attacks reported in Table \ref{tab:benchmarkattacks} on ViT-S/16 trained with DINO. True label is "Pelican". Last row indicates predicted class after the attack. Perturbations include important artifacts as $\epsilon$ grows.}
\label{tab:adversarialimages}
\end{table*}

\subsection{Adversarial Attacks}

Adversarial attacks are artificially crafted inputs designed to fool machine learning models \cite{szegedy2013intriguing}. Two main types of attacks can be found in the literature: white-box and black-box attacks. In white-box attacks, the attacker has full access to the model including its parameters and loss. The adversary may maximize the loss with respect to the target class for each input to achieve misclassification. Some attacks within this category are FGSM \cite{goodfellow2014explaining}, PGD \cite{madry2017towards} and C\&W \cite{carlini2017towards}. On the other hand, under black-box attacks, the adversary has no information about the model but can only observe inputs and corresponding predictions \cite{papernot2017practical,chen2017zoo}.

Previous work has shown supervised ViTs to be more robust than traditional convolutional classifiers against adversarial attacks \cite{aldahdooh2021reveal, shao2021adversarial}. Moreover, attacks crafted on supervised ViTs exhibit worse transferability across architectures \cite{naseer2022on}. 

Self-supervised ViTs trained with DINO have only been evaluated on transfer attacks for segmentation tasks \cite{naseer2022on}. However, their robustness have not been assessed for classification tasks nor compared with different architectures.

\subsubsection{Defenses against Adversarial Attacks}
Numerous defense strategies have been explored for different attacks and architectures in Computer Vision \cite{ren2020adversarial}. We group them into four main categories:

\textbf{Adversarial training} \cite{szegedy2013intriguing, madry2017towards}. When training the network, optimization becomes a $min$-$max$ problem. For each input $x$, we find the perturbation $x'$ that maximizes the loss in the $l_{p}$ ball of radius $\epsilon$ around $x$. Then, we try to find the optimal parameters ($\theta^*$) by minimizing the loss for $x'$ (see Equation \ref{eq:advtraining}). In practice, the search for $x'$ can be approximated efficiently with PGD \cite{madry2017towards}. Adversarial training increases resistance to attacks at the cost of accuracy on original samples and expensive optimization \cite{roth2019odds}. Subsequent work by Tramer et al. shows that this form of adversarial training converges to a degenerate global minimum. They propose \emph{Ensemble Adversarial Training} as an effective alternative that augments training data with attacks generated on different networks \cite{tramer2017ensemble}.

\vspace{-1.3em}
\begin{equation}
    \theta^* = \arg\min_{\theta} \max_{||r||_{p} \leq \epsilon} loss(f_{\theta}(x+r), y)
\label{eq:advtraining}
\end{equation}
\vspace{-1.5em}

\textbf{Preprocessing}. Preprocessing input images may remove perturbations induced by attackers and lead to more robust predictions \cite{guo2017countering, buckman2018thermometer, yang2019menet}. JPEG compression has been shown to be an efficient defense, but it is not sufficient as the strength of the attack increases \cite{dziugaite2016study,das2017keeping}.

\textbf{Post-hoc detectors}. Detect whether an input sample was adversarial \cite{feinman2017detecting}. For example, using output logits \cite{roth2019odds,mosca2022suspicious} or SHAP values \cite{fidel2020explainability}. These systems provide an alert if an attack is detected, but generally do not find the true class.

\textbf{Ensemble models}. Combine multiple models for a given task. It is motivated by the fact that an attack may fool one architecture but not every model in the ensemble. Abbasi et al.\cite{abbasi2017robustness} makes use of specialist classifiers trained on the classes most prone to misclassification.

However, if an attacker gets to know a defense, \emph{adaptitve attacks} may be crafted to surpass them \cite{carlini2017adversarial, tramer2017ensemble}.

\section{Adversarial Attacks on Self-Supervised ViT}
\label{sec:analysis}
\begin{table*}[h]
\centering
\begin{adjustbox}{max width=\textwidth}
\begin{tabular}{lc|ccc|ccc|c|c|}
\cline{3-10}
\multicolumn{1}{c}{}                        &          & \multicolumn{3}{c|}{\textbf{FGSM}}                                                                             & \multicolumn{3}{c|}{\textbf{PGD}}                                                                             & \textbf{C\&W} & \multirow{2}{*}{\textbf{Clean}} \\ \cline{3-9}
\multicolumn{1}{c}{}                        &          & \multicolumn{1}{c|}{$\epsilon = 0.001$} & \multicolumn{1}{c|}{$\epsilon = 0.03$} & {$\epsilon = 0.1$} & \multicolumn{1}{c|}{$\epsilon = 0.001$} & \multicolumn{1}{c|}{$\epsilon = 0.03$} & {$\epsilon = 0.1$} & {$c = 50$}    &                        \\ \hline
\multicolumn{1}{|l|}{\multirow{2}{*}{DINO}} & ViT-S/16 & \multicolumn{1}{c|}{52.4\%}             & \multicolumn{1}{c|}{0.9\%}             & 1.1\%              & \multicolumn{1}{c|}{49.6\%}             & \multicolumn{1}{c|}{0.0\%}             & 0.0\%              & 0.2\%         & 76.8\%                 \\ \cline{2-10} 
\multicolumn{1}{|l|}{}                      & ViT-B/16 & \multicolumn{1}{c|}{\textbf{58.9\%} }            & \multicolumn{1}{c|}{1.8\%}             & 1.5\%              & \multicolumn{1}{c|}{\textbf{56.8\%}   }          & \multicolumn{1}{c|}{0.0\%}             & 0.0\%              & 0.4\%         & 77.9\%                 \\ \hline
\multicolumn{1}{|l|}{Superv.}               & ViT-B/16 & \multicolumn{1}{c|}{55.1\%}             & \multicolumn{1}{c|}{\textbf{17.3\%} }           & 14.5\%   & \multicolumn{1}{c|}{47.7\%}             & \multicolumn{1}{c|}{\textbf{0.7\%}}             & \textbf{0.1}\%              & 0.8\%         & \textbf{80.2\%}        \\ \specialrule{.2em}{0em}{0em} 
\multicolumn{2}{|c|}{ResNet-50}                        & \multicolumn{1}{c|}{47.8\%}             & \multicolumn{1}{c|}{8.0\%}             & \textbf{24.3\%}             & \multicolumn{1}{c|}{43.9\%}             & \multicolumn{1}{c|}{0.1\%}             & 0.0\%              & \textbf{7.2}\%        & 75.7\%                 \\ \hline
\end{tabular}
\end{adjustbox}
\caption{Accuracy of ViT models trained using DINO and supervised (Superv.) learning against different white-box adversarial attacks. Metrics are computed on the whole ImageNet validation set. Last column represents accuracy on the original samples from which adversarial images were generated. ResNet-50 is included as a benchmark for convolutional architectures.}
\label{tab:benchmarkattacks}
\end{table*}

As motivated previously, the goal of this section is to evaluate whether features learnt through DINO are more robust against adversarial perturbations that preserve input similarity. Although adversarial accuracy of supervised Vision Transformers has already been assessed in previous works \cite{aldahdooh2021reveal, mahmood2021robustness}, there are no existing benchmarks for self-supervised ViTs trained using DINO. Moreover, a comparison between these two training schemes has not been done yet. In view of fairness, we produce results using the same Transformer architecture. We consider the supervised ViT-B/16 HuggingFace implementation \cite{wolf2019huggingface}, and its equivalent self-supervised ViT-B/16 taken from the DINO official repository\footnote{https://github.com/facebookresearch/dino}. Both have 85M parameters and are fine-tuned for classification on ImageNet-1k \cite{russakovsky2015imagenet}. We also include a smaller self-supervised ViT-S/16 (21M parameters) to measure robustness across different architecture sizes, and ResNet-50 \cite{he2016deep} as benchmark for CNNs.

We evaluate all four models on the same random split containing 3000 samples from ImageNet validation set. Their corresponding adversarial images are crafted using white-box attacks. We test against FGSM ($L_{\infty}$), PGD ($L_{\infty}$) and C\&W ($L_{2}$) using the implementation provided by \cite{kim2020torchattacks}. Attack parameters are chosen through visual exploration (see Table \ref{tab:adversarialimages}). $\epsilon=0.001$ is a subtle attack that does not introduce artifacts in the images. $\epsilon=0.03$ yields small artifacts, and finally, $\epsilon=0.1$ as an upperbound with pronounced perturbations. Accuracy on each of the attacks is reported in Table \ref{tab:benchmarkattacks}. Results show that ViTs trained using DINO do not exhibit a significant difference in terms of robustness to these white-box attacks compared to their supervised counterparts.

\begin{table}[h]
\resizebox{\linewidth}{!}{
\begin{tabular}{l|l|l|l|l|}
\cline{2-5}
                                & ViT-S (D) & ViT-B (D) & ViT-B (S) & ResNet-50 \\ \hline
\multicolumn{1}{|l|}{ViT-S (D)} & \cellcolor[HTML]{FFCCC9}0.0\% & 12.5\%                         & 41.1\%                        & 51.3\%                        \\ \hline
\multicolumn{1}{|l|}{ViT-B (D)} & 5.2\%                         & \cellcolor[HTML]{FFCCC9} 0.0\% & 31.4\%                        & 48.5\%                        \\ \hline
\multicolumn{1}{|l|}{ViT-B (S)} & 47.1\%                        & 47.8\%                         & \cellcolor[HTML]{FFCCC9}0.8\% & 59.7\%                         \\ \hline
\multicolumn{1}{|l|}{ResNet-50} & 65.2\%                        & 68.4\%                         & 74.9\%                        & \cellcolor[HTML]{FFCCC9}0.1\% \\ \hline
\end{tabular}}
\caption{Classification accuracy of adversarial samples transferred across architectures. All attacks were crafted using PGD ($\epsilon=0.03$). Rows represent generation setups and columns, the network used for evaluation. Computed on all validation images from ImageNet-1k. Patch size (16) has been omitted for clarity. (S) and (D) indicate Supervised and DINO training respectively.\label{tab:transfer}}
\end{table}

\begin{figure*}[h]
     \centering
     \begin{subfigure}{0.3\textwidth}
         \centering
         \includegraphics[width=\textwidth]{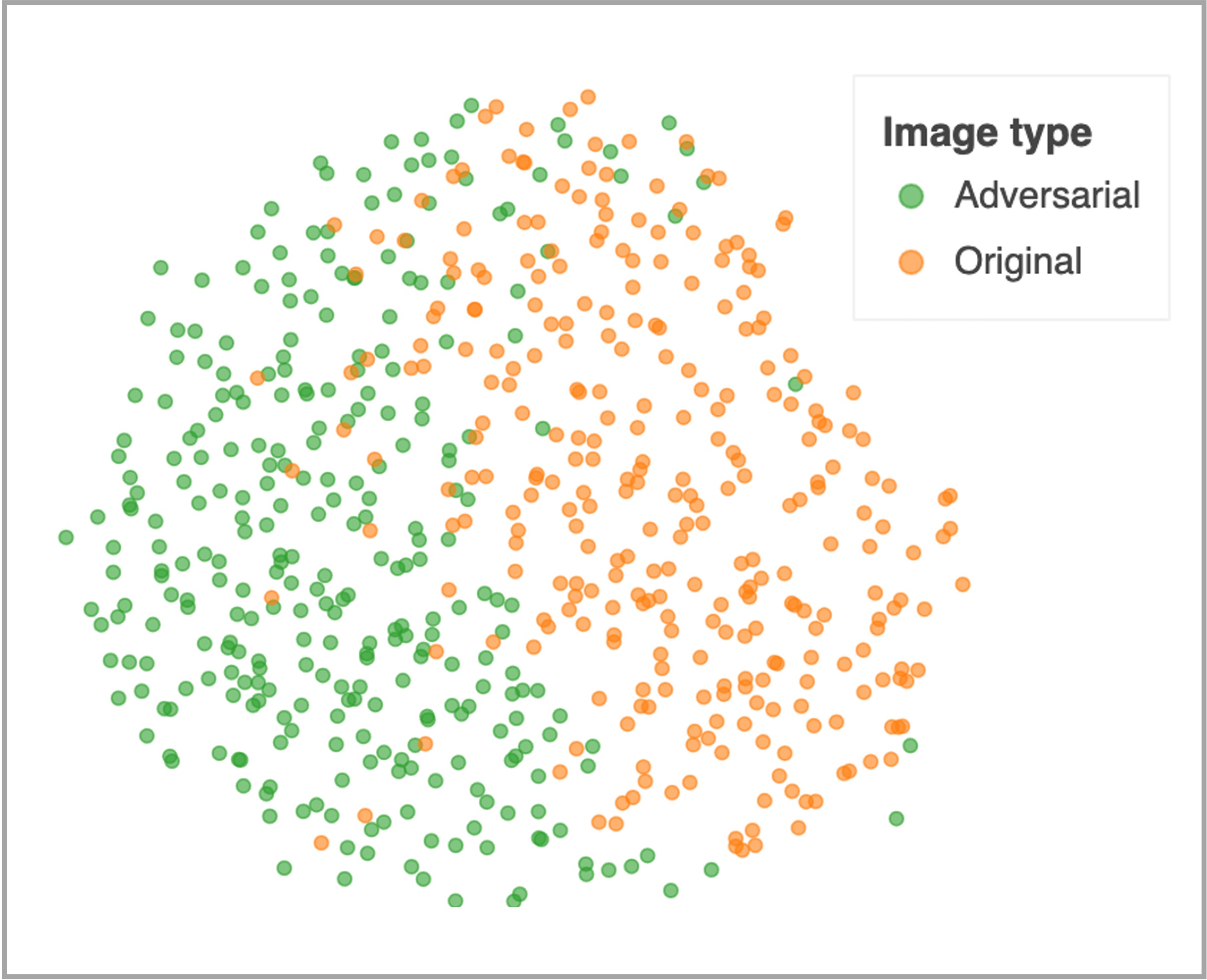}
         \caption{Random images and their adversarial perturbations. Color shows whether a samples is adversarial or original.}
         \label{fig:umap1}
     \end{subfigure}
     \hfill
     \begin{subfigure}{0.3\textwidth}
         \centering
         \includegraphics[width=\textwidth]{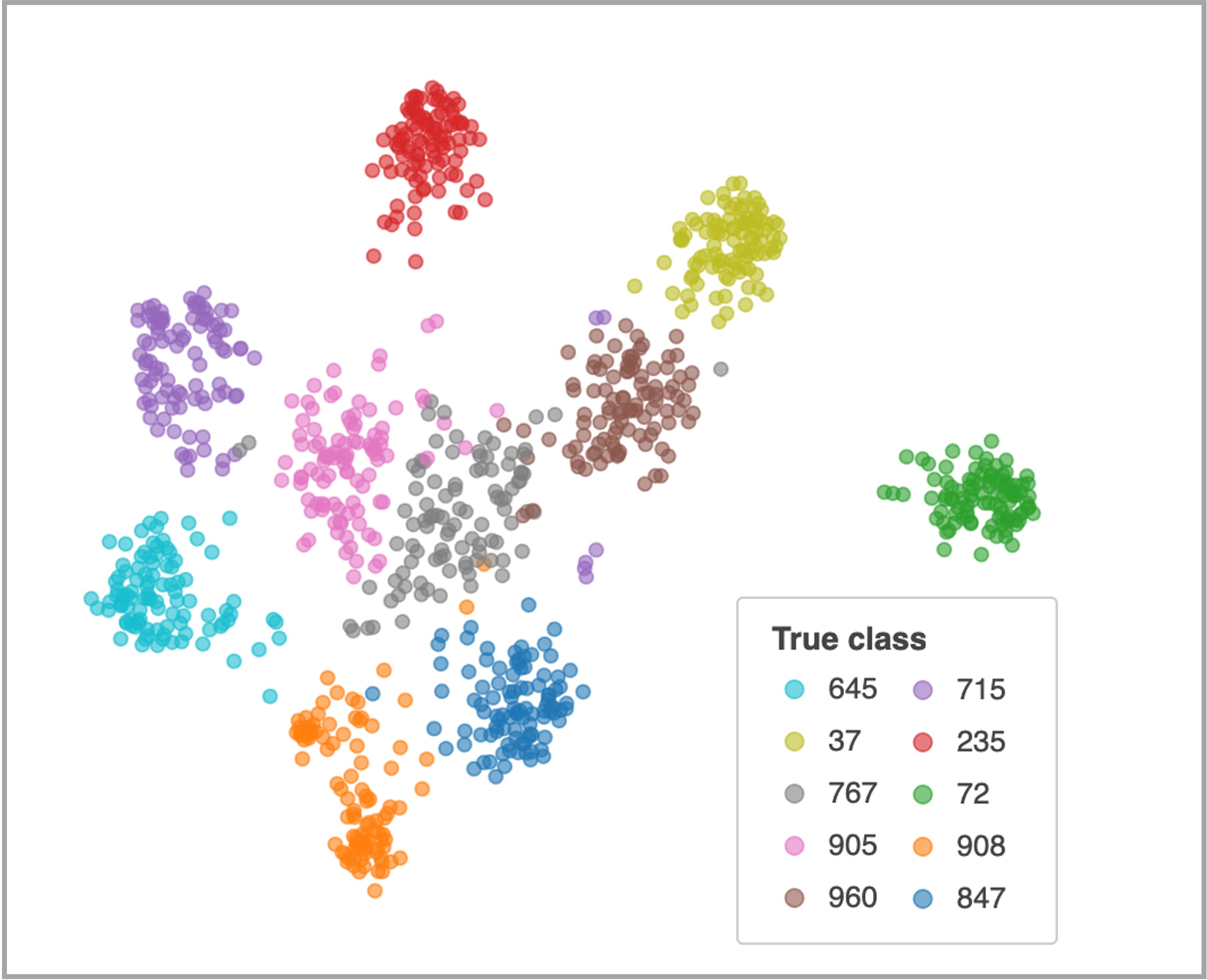}
         \caption{Original and adversarial images from 10 random classes. Color represents true label.}
         \label{fig:umap2}
     \end{subfigure}
     \hfill
     \begin{subfigure}{0.3\textwidth}
         \centering
         \includegraphics[width=\textwidth]{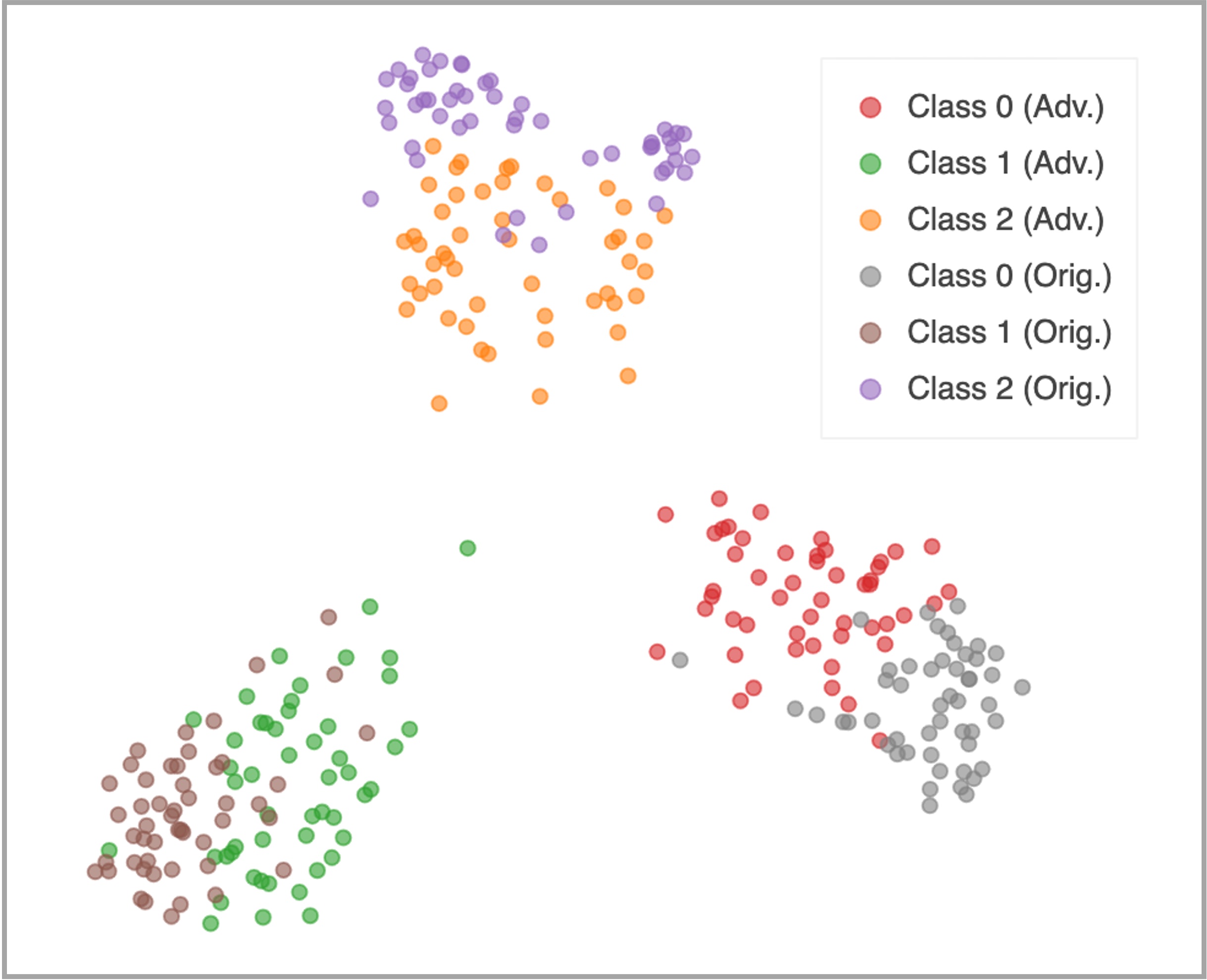}
         \caption{Adversarial and original images from 3 random classes. Color indicates true class, and if it is an attack.}
         \label{fig:umap3}
     \end{subfigure}
    \caption{UMAP visualization (\textit{canberra} distance) of the output space generated by ViT-S/16 on ImageNet validation images and their adversarial perturbations generated using PGD ($L_{\infty}$ and $\epsilon=0.03$).}
    \label{fig:umap}
\end{figure*}

\subsection{Adversarial Attacks Transferability}
Another well-known property from adversarial attacks is their transferability across architectures \cite{goodfellow2014explaining,szegedy2013intriguing,demontis2019adversarial}. We compare how attacks generated on supervised and self-supervised ViTs, and ResNet-50 transfer among them. Our results in Table \ref{tab:transfer} show that DINO ViTs are equally vulnerable to such adversaries, especially when another DINO ViT is used as source. Overall, ViTs are more robust against attacks generated on ResNet-50 than against those from transformers.

Our results reproduce the limited transferability between supervised ViTs and convolutional models presented by \cite{naseer2022on}. However, they also reveal that self-supervision increases transferability in both directions. 

\subsection{Latent Space for Adversarial Attacks}
\label{sec:latent_space_for_adversarial_attacks}
This section motivates how attacks affect DINO ViTs by analyzing the latent representation for adversarial inputs which would then be used for classification.

For the experiments, we use a ViT-S/16 trained using DINO. The latent space for this model, following its original implementation, is constructed by concatenating the \texttt{[CLS]} embedding from the last 4 layers of the ViT encoder. This results in a 1536 dimensional representation for each input image. The adversarial images are crafted using PGD with $L_{\infty}$ and $\epsilon=0.03$. As shown in Table \ref{tab:benchmarkattacks}, this attack has 100\% success rate in the selected architecture, i.e. all adversarial images are incorrectly classified by the model.

We use UMAP \cite{sainburg2021parametric} and different random samples of ImageNet validation set to interpret this high-dimensional space (see Figure \ref{fig:umap}). We identify three relevant properties. (i) Adversarial inputs can be separated from original images in this space (see Figure \ref{fig:umap1}). (ii) Adversarial images stay close to original samples in their true class, i.e., the class predicted before perturbations (see Figure \ref{fig:umap2}). (iii) Although adversarial inputs are close to original, they remain separable within the clusters (see Figure \ref{fig:umap3}).

These properties show that the latent space may comprise enough information to linearly separate adversarial samples without retraining the ViT. This is crucial for setups in which computational resources are limited and will motivate our \emph{Ensemble of Specialized Networks}.

\section{Adversarial Defense Strategies}
\label{sec:defenses}
We present three defense strategies for a \emph{target model} using a self-supervised DINO ViT as backbone, and optimized for classification on ImageNet \cite{deng2009imagenet}. Since training on the whole ImageNet dataset is known to be a computationally expensive problem, we limit ourselves to a restricted subset used in previous works \cite{tsipras2018robustness, engstrom2019adversarial}. It comprises semantically similar images into 9 super-classes (see Table \ref{tab:dataset}). The reduced dataset is then balanced so that all super-classes contain the same number of samples, resulting in 93,600 training and 3,600 validation images. 

All things considered, we create our \emph{target model} by training a linear classification layer on top of a ViT-S/16 encoder using the 9-class reduced dataset. It obtains 97.3\% classification accuracy in this task. We will use this architecture and reduced dataset throughout the remaining sections.

\subsection{Adversarial Training}
Assuming adversarial training on the whole Transformer architecture is a very costly process, our goal is to test whether it is possible to perform adversarial training just on the classification head while keeping the ViT frozen to reduce complexity.

In the foundational paper for adversarial training, it was already pointed out that capacity in the network is very important to achieve high accuracy \cite{madry2017towards}. We found out that a linear classification head didn't generalize properly, so we consider a fully connected neural network with 3 hidden layers (2048, 1024, 512) and ReLU activation.

\begin{figure}[h]
  \centering
  \includegraphics[width=0.4\textwidth]{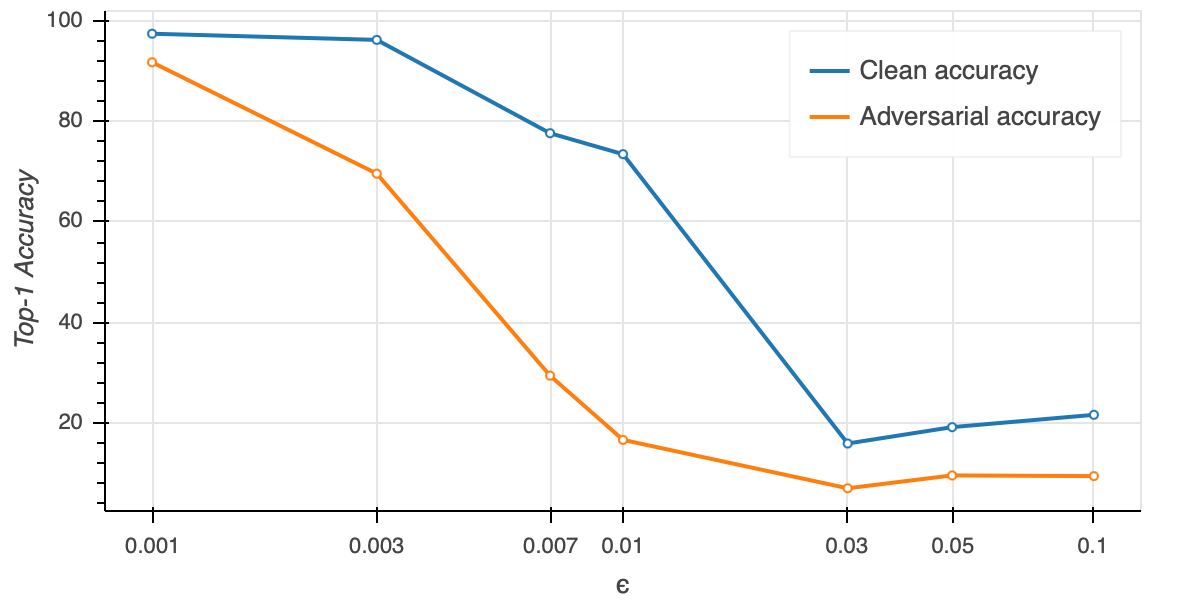}
  \caption{Linear-log plot for top-1 accuracy on validation set after performing PGD adversarial training for different values of $\epsilon$.}
  \label{fig:advtraining}
\end{figure}

We replicate adversarial training as presented by Madry et al. by optimizing Equation \ref{eq:advtraining}. Maximization of the loss is approximated with PGD since it has been shown to generalize best \cite{madry2017towards}. The attack is always crafted to fool the latest and most robust version of the classifier during training. Figure \ref{fig:advtraining} plots top-1 accuracy on clean and adversarial samples after training the classification head for 5 epochs. Adversarial Training can be achieved by only fine-tuning the classification head for low values of $\epsilon$, but collapses as it increases.

\begin{figure}[h]
     \centering
     \begin{subfigure}{0.21\textwidth}
         \centering
         \includegraphics[width=\textwidth]{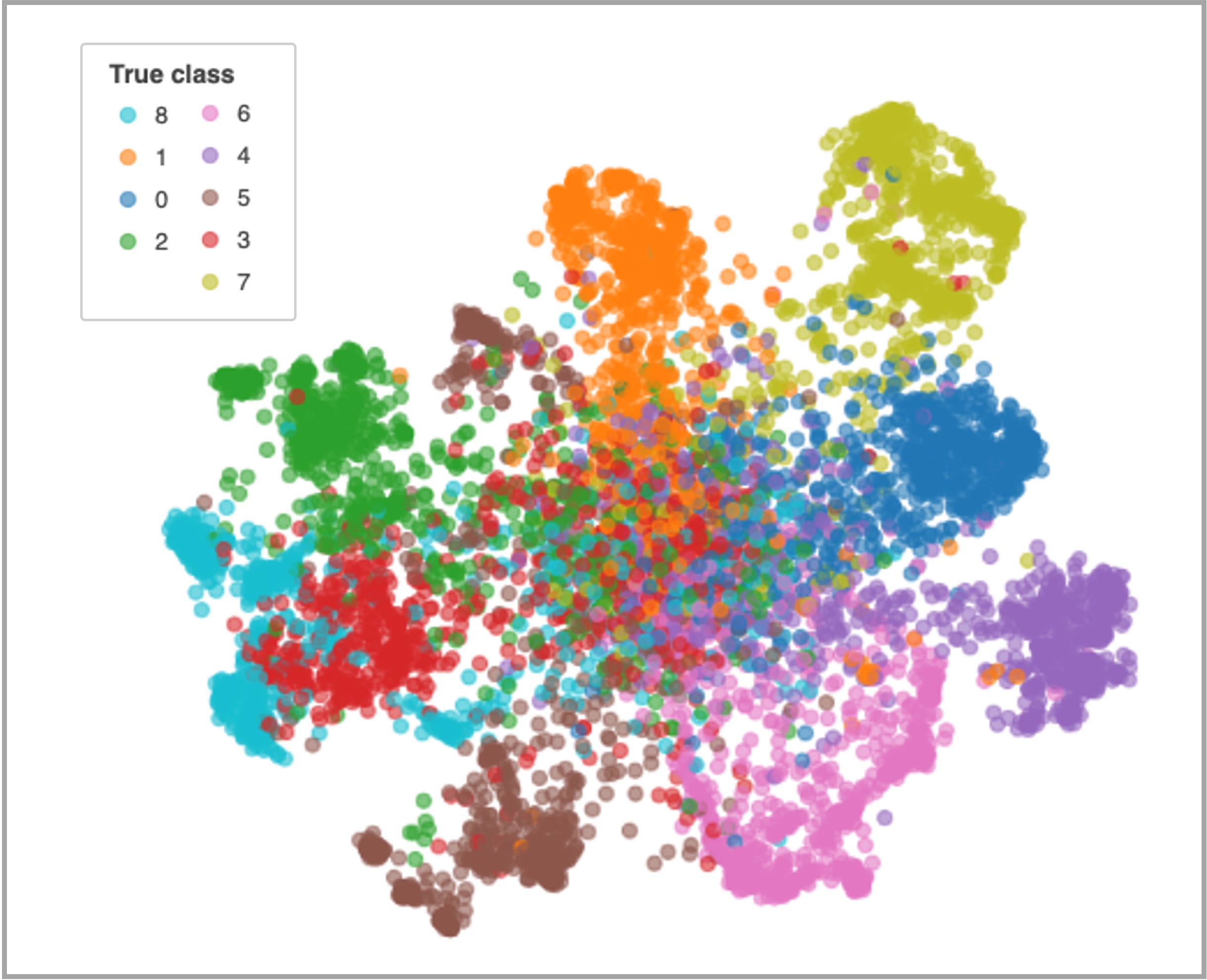}
         \caption{Original and adversarial images colored according to their true class.}
         \label{fig:umapadvtraining1}
     \end{subfigure}
     \hspace{0.02\textwidth}
     \begin{subfigure}{0.21\textwidth}
         \centering
         \includegraphics[width=\textwidth]{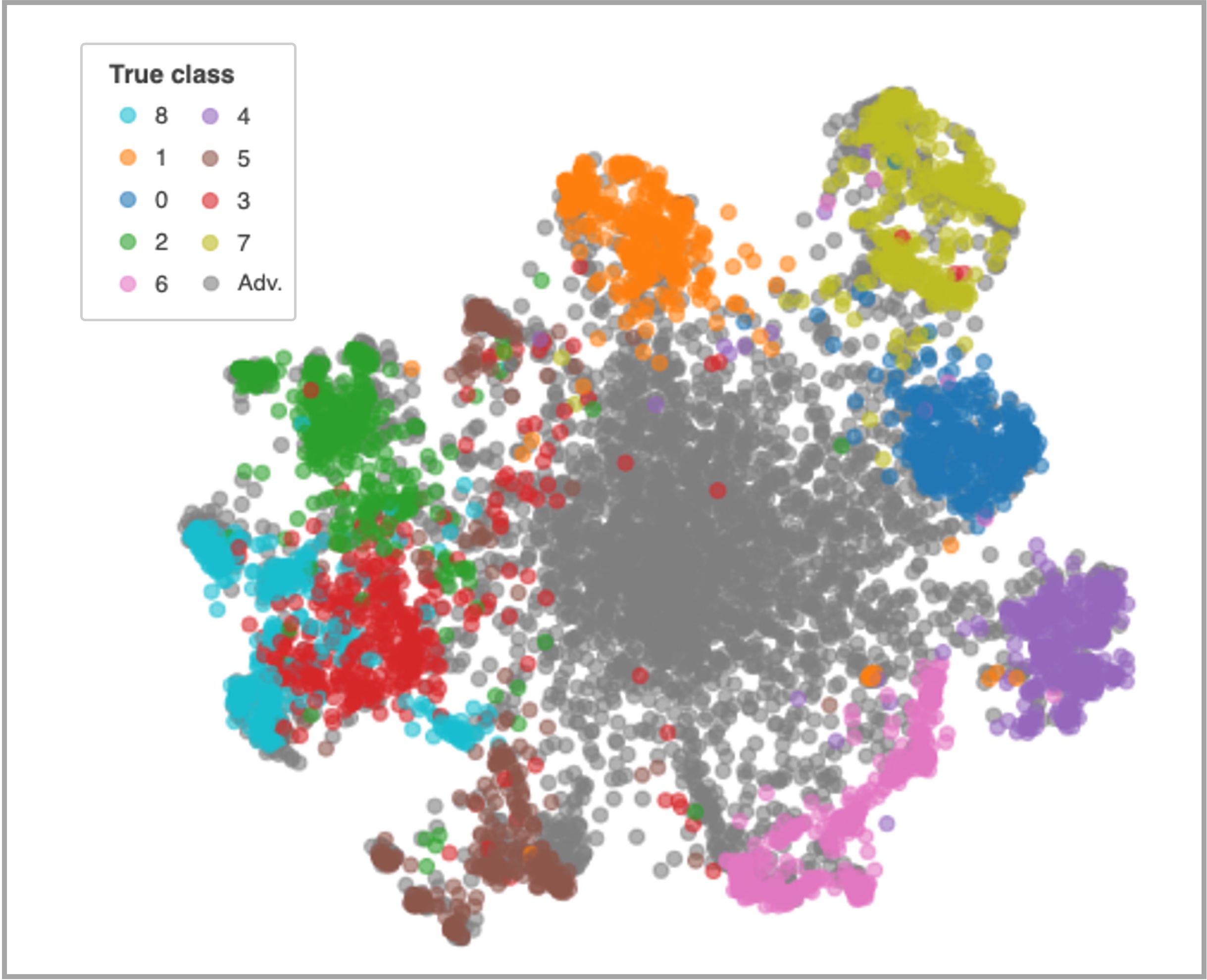}
         \caption{Original images colored according to their true class and adversarial in grey.}
         \label{fig:umapadvtraining2}
     \end{subfigure}
        \caption{UMAP visualization (\textit{canberra} distance) of training images and their adversarial perturbation using PGD ($L_{\infty}$ and $\epsilon=0.03$) during adversarial training. Figure (a) represents all images according to their true class and Figure (b) shows which of those points correspond to adversarial images.}
        \label{fig:umapadvtraining}
\end{figure}

Finally, we motivate this collapse. Figure \ref{fig:umapadvtraining} depicts the latent representation generated by the ViT for adversarial (PGD with $\epsilon = 0.03$) and clean samples. This space is then used as input for classification. It shows how the attack was able to generate samples that are mapped to a region which is no longer separable by the classifier. This is most likely due to the limitations that freezing the Transformer imposes on our representational power.
    
\subsection{Ensemble Adversarial Training}
\label{subsubsec:dataaugmentation}

\begin{table*}[h]
\centering
\begin{tabular}{cc|c|ccc|ccc|c|}
\cline{3-10}
                                  &            & \multirow{2}{*}{\textbf{Clean}} & \multicolumn{3}{c|}{\textbf{PGD}}                                                                            & \multicolumn{3}{c|}{\textbf{FGSM}}                                                                  & \textbf{\textbf{C\&W}} \\ \cline{1-2} \cline{4-10} 
\multicolumn{1}{|c|}{\textbf{Strategy}}    & $\epsilon$ &                        & \multicolumn{1}{c|}{$\epsilon = 0.001$} & \multicolumn{1}{c|}{$\epsilon = 0.03$} & $\epsilon = 0.1$ & \multicolumn{1}{c|}{$\epsilon = 0.001$} & \multicolumn{1}{c|}{$\epsilon = 0.03$} & $\epsilon = 0.1$ & $c = 50$               \\ \specialrule{.2em}{0em}{0em} 
\multicolumn{1}{|c|}{No defense}  & -          & 97.3\%                 & \multicolumn{1}{c|}{89.2\%}             & \multicolumn{1}{c|}{0.0\%}             & 0.0\%            & \multicolumn{1}{c|}{90.3\%}                   & \multicolumn{1}{c|}{1.5\%}                  & 3.1\%                 &           0.1\%             \\ \specialrule{.2em}{0em}{0em} 
\multicolumn{1}{|c|}{Ensemble AT} & 0.001      & 97.3\%                 & \multicolumn{1}{c|}{92.6\%}             & \multicolumn{1}{c|}{0.3\%}             & 0.1\%            & \multicolumn{1}{c|}{93.2\%}             & \multicolumn{1}{c|}{13.8\%}            & 12.3\%           & 55.4\%                 \\ \hline
\multicolumn{1}{|c|}{Ensemble AT} & 0.03       & 93.3\%                 & \multicolumn{1}{c|}{91.6\%}             & \multicolumn{1}{c|}{98.7\%}            & 97.5\%           & \multicolumn{1}{c|}{91.9\%}             & \multicolumn{1}{c|}{\textbf{89.8\%}}            & \textbf{76.1\% }          & \textbf{86.6\%}                 \\ \hline
\multicolumn{1}{|c|}{Ensemble AT} & 0.1        & 95.3\%                 & \multicolumn{1}{c|}{91.9\%}             & \multicolumn{1}{c|}{90.1\%}            & 98.7\%           & \multicolumn{1}{c|}{91.9\%}             & \multicolumn{1}{c|}{85.5\%}            & 75.4\%           & 80.0\%                 \\ \specialrule{.2em}{0em}{0em} 
\multicolumn{1}{|c|}{Specialized Net.}    & 0.001      & 96.9\%                 & \multicolumn{1}{c|}{\textbf{92.9\%}}             & \multicolumn{1}{c|}{1.1\%}             & 1.2\%            & \multicolumn{1}{c|}{\textbf{93.3\%}}             & \multicolumn{1}{c|}{26.6\%}            & 20.6\%           & 71.0\%                 \\ \hline
\multicolumn{1}{|c|}{Specialized Net.}    & 0.03       & 96.9\%                 & \multicolumn{1}{c|}{88.6\%}             & \multicolumn{1}{c|}{\textbf{99.5\%}}            & 99.2\%           & \multicolumn{1}{c|}{89.7\%}             & \multicolumn{1}{c|}{84.3\%}            & 70.7\%           & 5.7\%                  \\ \hline
\multicolumn{1}{|c|}{Specialized Net.}    & 0.1        & \textbf{97.4\%}                & \multicolumn{1}{c|}{89.2\%}             & \multicolumn{1}{c|}{62.6\%}            & \textbf{99.6\%}           & \multicolumn{1}{c|}{90.3\%}             & \multicolumn{1}{c|}{72.1\%}            & 66.7\%           & 0.4\%                  \\ \hline
\end{tabular}
\caption{Accuracy for Ensemble Adversarial Training (AT) and Ensemble of Specialized Networks compared to a baseline classifier trained on the (reduced) ImageNet dataset. All defenses are built using PGD with the reported values for $\epsilon$. Columns indicate evaluation black-box attacks crafted on the \emph{target model} without defense.}
\label{tab:evaluation}
\end{table*}

As Tramer et al. showed in previous work, vanilla adversarial training may converge to a degenerate global minimum. Thus, they propose Ensemble Adversarial Training which consists in augmenting training data with perturbations transferred from a different network to decouple adversarial generation \cite{tramer2017ensemble}. In our setup, we consider attacks crafted on a surrogate of the \emph{target model} solving the same task. Thus, our robust classification layer will be optimized on an augmented dataset containing both clean and adversarial images generated on a static surrogate; i.e. the model generating adversarial samples does not change as training progresses. The main difference with vanilla adversarial training is that attacks are no longer crafted to fool the last iteration of the trained classifier. For each batch, we uniformly select between clean images or their adversarial counterparts. We use PGD to generate the augmented data using the surrogate.

Table \ref{tab:evaluation} (rows 2-4) reports the performance obtained by the resulting classifiers on attacks crafted on the \emph{target model} and validation images. Overall, the model trained on data augmented with PGD and $\epsilon=0.03$ exhibits the most robust generalization to attackers of different natures and strengths, with an average accuracy of 87.3\%. This comes at the cost of only 4\% accuracy on clean samples. Then, we also test whether these defenses are robust against attacks crafted on a larger ViT/B-16. Whereas the \emph{target model} achieved 7.7\% on such attack, Ensemble Adversarial Training results in a model which correctly classifies 66\% of the adversarial samples (see rows 1-4 in Table \ref{tab:transferevaulation}). We leave as future work how combining perturbations coming from varied networks and attacks can improve generalization.

\begin{table}[]
\centering
\begin{tabular}{|c|c|c|c|}
\hline
\textbf{Strategy}    & $\epsilon$ & \textbf{Accuracy} \\ \specialrule{.2em}{0em}{0em} 
No defense  & -          &     7.7\%      \\ \specialrule{.2em}{0em}{0em} 
Ensemble AT & 0.001     & 13.7\%   \\ \hline
Ensemble AT  & 0.03      & \textbf{66.0\%}   \\ \hline
Ensemble AT  & 0.1          & 51.6\%   \\ \specialrule{.2em}{0em}{0em} 
Specialized Net.    & 0.001        & 25.6\%   \\ \hline
Specialized Net.    & 0.03        & 43.3\%   \\ \hline
Specialized Net.    & 0.1       & 18.1\%   \\ \hline
\end{tabular}
\caption{Accuracy for Ensemble Adversarial Training (AT) and Ensemble of Specialized Networks trained on different values of $\epsilon$ against transfer attacks crafted on DINO ViT/B-16 using PGD ($\epsilon=0.03$).\label{tab:transferevaulation} \vspace{-2em}}
\end{table}

\subsection{Ensemble of Specialized Networks}
\label{sec:ensemble}

Drawing from Section \ref{sec:latent_space_for_adversarial_attacks}, we motivate a potential defense that leverages the separability of adversarial inputs in the latent space generated by self-supervised DINO Vision Transformers. As a first step, we build a \emph{post-hoc detector} that determines whether a given sample was perturbed by an adversary or not. As input, it takes the image latent representation generated by the ViT encoder, i.e. the same input as the classification head. We use a linear layer that maps this 1536 dimensional vector into 2 classes: adversarial and original. Then, we train an \emph{adversarial classifier} specialized in correctly labelling perturbed inputs.

These two steps can be achieved with high accuracy, validating the property of adversarial inputs being separable in the latent space introduced in Section \ref{sec:analysis}. Although we just report end-to-end performance of the ensemble, detailed intermediate results can be found in Appendix \ref{sec:posthocappendix}.

Finally, since both clean and adversarial inputs can be precisely classified by their respective \emph{specialized classification heads}, we ensemble them by means of the post-hoc detector to build a robust end-to-end model. At inference time, we use the adversarial classifier whenever the post-hoc detector outputs more than 50\% probability of a sample being adversarial, and the original classifier trained on (reduced) ImageNet otherwise. We evaluate the performance of the ensemble pipeline on adversarial samples generated with various attacks on the \emph{target model} (ViT/S-16), as well as on ViT-B/16 to test against transfer attacks. The last three rows in Table \ref{tab:evaluation} report the performance of ensembles using a post-hoc detector and adversarial classifier trained on the same attack. However, an extensive analysis of all possible combinations is included in the Appendix (see Table \ref{table:ensemble_validation}). Accuracy of over 70\% is obtained for all attacks generated using PGD and FGSM, but drops for C\&W when compared with Ensemble Adversarial Training. This accuracy across the attacks comes at a minimal cost on the accuracy for the unperturbed samples (decreasing at most 0.4\%). Lastly, although the ensemble improves on the clean classifier for samples perturbed on ViT/B-16, its performance falls behind Ensemble Adversarial Training (see last 3 rows in Table \ref{tab:transferevaulation})


\section{Conclusion}




We have shown that features emerging from self-supervision in Vision Transformers using DINO do not bear a significant difference in terms of robustness to adversarial attacks compared to their equivalent supervised architectures and CNNs. Our results also reveal that self-supervision increases transferability of attacks between ViTs and CNNs. Furthermore, our analysis of the latent space generated by the Transformer provides the first insights into how such features are perturbed by adversarial attacks. 

Building on these insights, we evaluate the suitability of the latent space for different defense strategies that do not require retraining the ViT backbone. This is especially relevant for scenarios with limited computational resources. First, we showed that vanilla Adversarial Training can be successful for small perturbations, but may require unfreezing of the Transformer with increasing strength of the attack. Then, we used Ensemble Adversarial Training to augment training data with perturbations transferred from a surrogate. Finally, we built an ensemble network in which a post-hoc detector determines if an input is adversarial and according to its prediction, a specialized head for clean or adversarial inputs is used for classification.

Of these strategies, we report that Ensemble Adversarial Training results in the most robust behaviour across different attackers and at a small cost on accuracy for clean samples. It is able to increase average accuracy against adversaries from 24.0\% to 87.3\% at the cost of only 4\% accuracy on clean samples.

\section{Limitations \& Future Work}

\textbf{Limited Dataset.} Explore how the performance of such defenses generalize to the entire ImageNet.

\textbf{Adversarial training}. It remains unexplored if unfreezing some layers in the encoder could lead to successful AT for larger perturbations.

\textbf{Ensemble Adversarial Training}. Analyze how to optimally augment training data to increase generalization.

\textbf{Further self-supervision strategies}. Assess if these properties and defenses generalize to ViTs trained using other self-supervision strategies.

\textbf{Adaptive attacks}. Explore if our defenses can be surpassed by crafting a specific adaptive attack.

\textbf{Latent space.} We build the latent space as presented in the original DINO work. Future work may explore the impact of incorporating information from more layers.

\section*{Acknowledgements}

We thank our colleagues from the Data Analytics Lab at ETH Zurich who provided us with the compute power and valuable insights on the work, and reviewers for their contributions to improve our work. 

We also want to express our gratitude to Luis Salamanca for his initial motivation to work with DINO, and Anna Susmelj for her thoughtful feedback.

\bibliography{main}
\bibliographystyle{icml2022}

\newpage
\appendix
\onecolumn

\section{Setup}
All the experiments presented in this paper have been performed using the Euler cluster at ETH Zürich. More specifically, our working environment used 4 CPU cores with 10240MB of RAM each, and 1 GPU (GeForceRTX2080Ti).

\section{Dataset}
For our experiments we limit ourselves to a restricted dataset considered in previous works \cite{tsipras2018robustness, engstrom2019adversarial}. It comprises 311 semantically similar classes into 9 super-classes. The reduced dataset is then balanced so that all super-classes contain the same number of images. This results in 93,600 training images and 3600 for validation. The following table displays each of the resulting super-classes and their child classes in the original ImageNet Dataset. Library by Engstrom et al. was used to create the reduced dataset \cite{robustness2019engstrom}.

\begin{table}[h]
    \centering
    \begin{tabular}{|c|c|}
    \hline
        \textbf{Super-class} & \textbf{ImageNet classes (inclusive)} \\ \hline
        Dog & 151 to 268 \\ \hline
        Cat & 281 to 285 \\ \hline
        Frog & 30 to 32 \\ \hline
        Turtle & 33 to 37 \\ \hline
        Bird & 80 to 100 \\ \hline
        Primate & 365 to 382 \\ \hline
        Fish & 389 to 397 \\ \hline
        Crab & 118 to 121 \\ \hline
        Insect & 300 to 319 \\ \hline
    \end{tabular}
    \caption{Classes from ImageNet-1k used to generate our restricted dataset.}
    \label{tab:dataset}
\end{table}

\section{Ensemble detailed metrics}
\label{sec:posthocappendix}
\begin{table*}[h]
\centering
\begin{adjustbox}{max width=\textwidth}
\begin{tabular}{l|ccc|ccc|c|}
\cline{2-8}
\multicolumn{1}{c|}{} & \multicolumn{3}{c|}{\textbf{PGD}} & \multicolumn{3}{c|}{\textbf{FGSM}} & \textbf{C\&W} \\ \cline{2-8}
\multicolumn{1}{c|}{} & \multicolumn{1}{c|}{$\epsilon = 0.001$} & \multicolumn{1}{c|}{$\epsilon = 0.03$} & \multicolumn{1}{l|}{$\epsilon = 0.1$} & \multicolumn{1}{c|}{$\epsilon = 0.001$} & \multicolumn{1}{c|}{$\epsilon = 0.03$} & \multicolumn{1}{l|}{$\epsilon = 0.1$} & $c = 50$      \\ \hline

\multicolumn{1}{|l|}{$\epsilon = 0.001$}  & \multicolumn{1}{c|}{58.2\%} & \multicolumn{1}{c|}{73.375\%} & 74.9\% & \multicolumn{1}{c|}{59.0\%} & \multicolumn{1}{c|}{75.708\%} & 75.8\% & 72.5\% \\ \hline

\multicolumn{1}{|l|}{$\epsilon = 0.03$}  &  \multicolumn{1}{c|}{50.2\%} & \multicolumn{1}{c|}{99.5\%} & 99.6\% & \multicolumn{1}{c|}{50.2\%} & \multicolumn{1}{c|}{99.6\%} & 99.7\% & 56.5\% \\ \hline

\multicolumn{1}{|l|}{$\epsilon = 0.1$} &  \multicolumn{1}{c|}{50.0\%} & \multicolumn{1}{c|}{81.4\%} & 99.9\% & \multicolumn{1}{c|}{50.0\%} & \multicolumn{1}{c|}{96.3\%} & 99.9\% & 50.3\% \\ \hline

\end{tabular}
\end{adjustbox}
\caption{Post-hoc detection accuracy. Rows represent $\epsilon$ used in PGD for training. The columns show  the different datasets used for evaluation.}
\label{tab:posthocmatrix_full}
\end{table*}

\begin{table}[h]
    \centering
    \begin{tabular}{ |l|c|c|c|c|} 
     \hline
      & $\epsilon = 0.001$& $\epsilon = 0.03$ & $\epsilon = 0.1$ & \textbf{Clean} \\
     \hline
     $\epsilon = 0.001$ & \textbf{93.58}\% & 1.08\% & 1.22\% & 96.78\%\\
     \hline
     $\epsilon = 0.03$ & 2.22\% & \textbf{99.86}\% & 99.25\% & 0.61\%\\
     \hline
     $\epsilon = 0.1$ & 1.75\% & 99.50\% & \textbf{99.64\%} & 0.31\%\\ 
     \hline
     Clean & 89.17\% & 0.00\% & 0.08\% & \textbf{97.53\%}\\ 
     \hline
    \end{tabular}
    \caption{Accuracy for different linear classifiers trained on adversarial datasets. All adversarial datasets were generated using PGD, where the $\epsilon$ was varied. Rows show the data used for training the classifier and columns represent the datasets used for evaluation.}
    \label{table:advlinearclassifier}
\end{table}

\begin{table*}[]
    \centering
    \begin{adjustbox}{max width=\textwidth}
        \begin{tabular}{|l|l|c|c|c|c|c|c|c|c|}
        \cline{1-10} \textbf{Post-Hoc} & \textbf{ImageNet} & \multirow{2}{*}{\textbf{Clean}} & \textbf{PGD} & \textbf{PGD} & \textbf{PGD} & \textbf{FGSM} & \textbf{FGSM} & \textbf{FGSM} & \textbf{C\&W}\\ 
        \cline{4-10} \textbf{Classifier} & \textbf{Classifier} & & \multicolumn{1}{c|}{$\epsilon = 0.001$} & $\epsilon = 0.03$ & $\epsilon = 0.1$ & $\epsilon = 0.001$ & $\epsilon = 0.03$ & $\epsilon = 0.1$ & 50 \\
        \hline
        \multicolumn{1}{|c|}{$\epsilon = 0.001$} & \multicolumn{1}{c|}{$\epsilon = 0.001$} & 96.89\% & 92.94\% & \multicolumn{1}{c|}{1.08\%} & 1.22\% & 93.33\% & 26.58\% & 20.61\% & 71.03\% \\ 
        \hline
        \multicolumn{1}{|c|}{$\epsilon = 0.001$} & \multicolumn{1}{c|}{$\epsilon = 0.03$} & 51.81\% & 35.58\% & \multicolumn{1}{c|}{94.89\%} & 97.39\% & 34.11\% & 84.08\% & 70.67\% & 29.36\%\\
        \hline
        \multicolumn{1}{|c|}{$\epsilon = 0.001$} & \multicolumn{1}{c|}{$\epsilon = 0.1$} & 51.64\% & 35.08\% & \multicolumn{1}{c|}{94.56\%} & 97.78\% & 33.67\% & 76.83\% & 66.67\% & 16.89\% \\
        \hline
        \multicolumn{1}{|c|}{$\epsilon = 0.03$} & \multicolumn{1}{c|}{$\epsilon = 0.001$} & 97.58\% & 89.36\% & \multicolumn{1}{c|}{1.08\%} & 1.22\% & 90.47\% & 26.58\% & 20.61\% & 9.91\% \\ 
        \hline
        \multicolumn{1}{|c|}{$\epsilon = 0.03$} & \multicolumn{1}{c|}{$\epsilon = 0.03$} & 96.92\% & 88.58\% & \multicolumn{1}{c|}{99.50\%} & 99.19\% & 89.67\% & 84.28\% & 70.67\% & 5.67\% \\
        \hline
        \multicolumn{1}{|c|}{$\epsilon = 0.03$} & \multicolumn{1}{c|}{$\epsilon = 0.1$} & 96.92\% & 88.53\% & \multicolumn{1}{c|}{99.14\%} & 99.58\% & 89.64\% & 76.94\% &  66.67\% & 3.83\% \\
        \hline
        \multicolumn{1}{|c|}{$\epsilon = 0.1$} & \multicolumn{1}{c|}{$\epsilon = 0.001$} & 97.53\% & 89.22\% & \multicolumn{1}{c|}{0.89\%} & 1.22\% & 90.36\% & 23.72\% & 20.61\% &  0.64\% \\ 
        \hline
        \multicolumn{1}{|c|}{$\epsilon = 0.1$} & \multicolumn{1}{c|}{$\epsilon = 0.03$} & 97.42\% & 89.17\% & \multicolumn{1}{c|}{62.83\%} & 99.22\% & 90.31\% & 78.33\% & 70.67\% & 0.50\% \\
        \hline
        \multicolumn{1}{|c|}{$\epsilon = 0.1$} & \multicolumn{1}{c|}{$\epsilon = 0.1$} & 97.42\% & 89.17\% & \multicolumn{1}{c|}{62.58\%} & 99.61\% & 90.31\% & 72.14\% & 66.67\% & 0.42 \%\\ 
        \hline
        \end{tabular}
    \end{adjustbox}
    \caption{Accuracy for all possible combinations of post-hoc classifiers and adversarial classification heads to build an Ensemble. They are evaluated on all available adversarial attacks.}
    \label{table:ensemble_validation}
\end{table*}

\begin{table*}[]
    \centering
    \begin{adjustbox}{max width=\textwidth}
        \begin{tabular}{|c|c|c|}
\hline
\multicolumn{1}{|l|}{\textbf{Post-Hoc Classifier}} & \multicolumn{1}{l|}{\textbf{ImageNet Classifier}} & \textbf{ViT/B-16} \\ \hline
$\epsilon = 0.001$                                 & $\epsilon = 0.001$                                & 25.61\%           \\ \hline
$\epsilon = 0.001$                                 & $\epsilon = 0.03$                                 & 44.08\%           \\ \hline
$\epsilon = 0.001$                                 & $\epsilon = 0.1$                                  & 34.92\%           \\ \hline
$\epsilon = 0.03$                                  & $\epsilon = 0.001$                                & 24.06\%           \\ \hline
$\epsilon = 0.03$                                  & $\epsilon = 0.03$                                 & 43.28\%           \\ \hline
$\epsilon = 0.03$                                  & $\epsilon = 0.1$                                  & 34.69\%           \\ \hline
$\epsilon = 0.1$                                   & $\epsilon = 0.001$                                & 11.72\%           \\ \hline
$\epsilon = 0.1$                                   & $\epsilon = 0.03$                                 & 19.42\%           \\ \hline
$\epsilon = 0.1$                                   & $\epsilon = 0.1$                                  & 18.14\%           \\ \hline
\end{tabular}
    \end{adjustbox}
    \caption{Accuracy against attacks crafted on ViT/B-16 using PGD ($\epsilon=0.03$) for all possible Ensemble models.}
    \label{table:ensemble_transfer}
\end{table*}

\end{document}